\documentclass{article}

\usepackage{PRIMEarxiv}

\usepackage[utf8]{inputenc} 
\usepackage[T1]{fontenc}    
\usepackage{hyperref}      
\usepackage{url}          
\usepackage{booktabs}      
\usepackage{amsfonts}      
\usepackage{nicefrac}       
\usepackage{microtype}     
\usepackage{lipsum}
\usepackage{fancyhdr}       
\usepackage{graphicx}       
\graphicspath{{media/}}     
\usepackage{amsthm}
\usepackage{caption}
\usepackage{tabularx}
\usepackage{float}

\usepackage{amsmath}
\usepackage{amsthm}
\usepackage{booktabs}
\usepackage[switch]{lineno}
\usepackage{amssymb}
\usepackage{amsthm}
\usepackage{bm}
\usepackage{color}

\usepackage{graphicx}
\usepackage{multirow} 
\usepackage{rotating} 
\usepackage{xcolor}  
\usepackage{enumitem}

\usepackage{adjustbox}
\usepackage{threeparttable}

\usepackage[ruled,linesnumbered]{algorithm2e}

\title{Machine Learning Algorithm for Noise Reduction and Disease-Causing Gene Feature Extraction in Gene Sequencing Data}

\author{
\textbf{Weichen Si}$^{1,2}$\thanks{Corresponding Weichen Si \texttt{weichensi14@outlook.com}} \hspace{2em}
\textbf{Yihao Ou}$^3$ \hspace{2em}
\textbf{Zhen Tian}$^4$ \\[0.5em]
$^1$Health Management Center, Fengtai District Education Committee of Beijing Municipality, China \\
$^2$Department of Traditional Chinese Medicine, Beijing Direction Community Health Service Station, China \\
$^3$Georgia Institute of Technology, USA \\
$^4$University of Glasgow, U.K. \\
\texttt{weichensi14@outlook.com, ouyihaoo@gmail.com, 2620920Z@student.gla.ac.uk}
}

\begin{document}
\maketitle

\begin{abstract}
In this study, we propose a machine learning-based method for noise reduction and disease-causing gene feature extraction in gene sequencing DeepSeqDenoise algorithm combines CNN and RNN to effectively remove the sequencing noise, and improves the signal-to-noise ratio by 9.4 dB. We screened 17 key features by feature engineering, and constructed an integrated learning model to predict disease-causing genes with 94.3\% accuracy. We successfully identified 57 new candidate disease-causing genes in a cardiovascular disease cohort validation, and detected 3 missed variants in clinical applications. The method significantly outperforms existing tools and provides strong support for accurate diagnosis of genetic diseases.
\end{abstract}


\section{Introduction}
The rapid development of gene sequencing technology has provided unprecedented opportunities for genetic disease research, but noise in sequencing data seriously affects the accuracy of downstream analyses and the identification of pathogenic variants \cite{1abhadionmhen2024machine, 2chafai2024emerging}. Most of the existing noise reduction methods are based on simple statistical models or quality threshold screening, with limited ability to identify complex error patterns \cite{3bi2021detecting, 4wei2023research}. Traditional pathogenic gene prediction tools such as CADD and SIFT usually focus on a single feature dimension and lack systematic consideration of multi-dimensional feature integration, resulting in insufficient accuracy in rare variants and low-coverage regions \cite{5das2025ensemble}. Similar strategies have also been applied in biomedical fields such as drug discovery, where computational models extract key features from drug ligands to discover novel candidates through simulation \cite{6wang2018stimulation, 7wang2019ellagic, 8yang2020mechanism}. In this study, we propose a framework for integrating deep learning for noise reduction in gene sequencing data and pathogenic gene feature extraction. The DeepSeqDenoise algorithm adopts a dual-encoder structure to efficiently capture sequencing error patterns; the feature engineering filters the key features from multi-dimensional dimensions, and constructs an integrated learning model for pathogenic gene prediction, which successfully identifies new candidate pathogenic genes in the cohort of cardiovascular diseases. The method provides algorithmic support for precise diagnosis of genetic diseases, which is valuable for promoting individualised healthcare and can be widely applied to a variety of genetic disease research and clinical diagnosis scenarios.

\section{Gene Sequencing Data Preprocessing and Noise Reduction}

\subsection{Sequencing Data Quality Control}
\begin{figure*}[!htbp]
	\centering
	\includegraphics[width=0.8\linewidth]{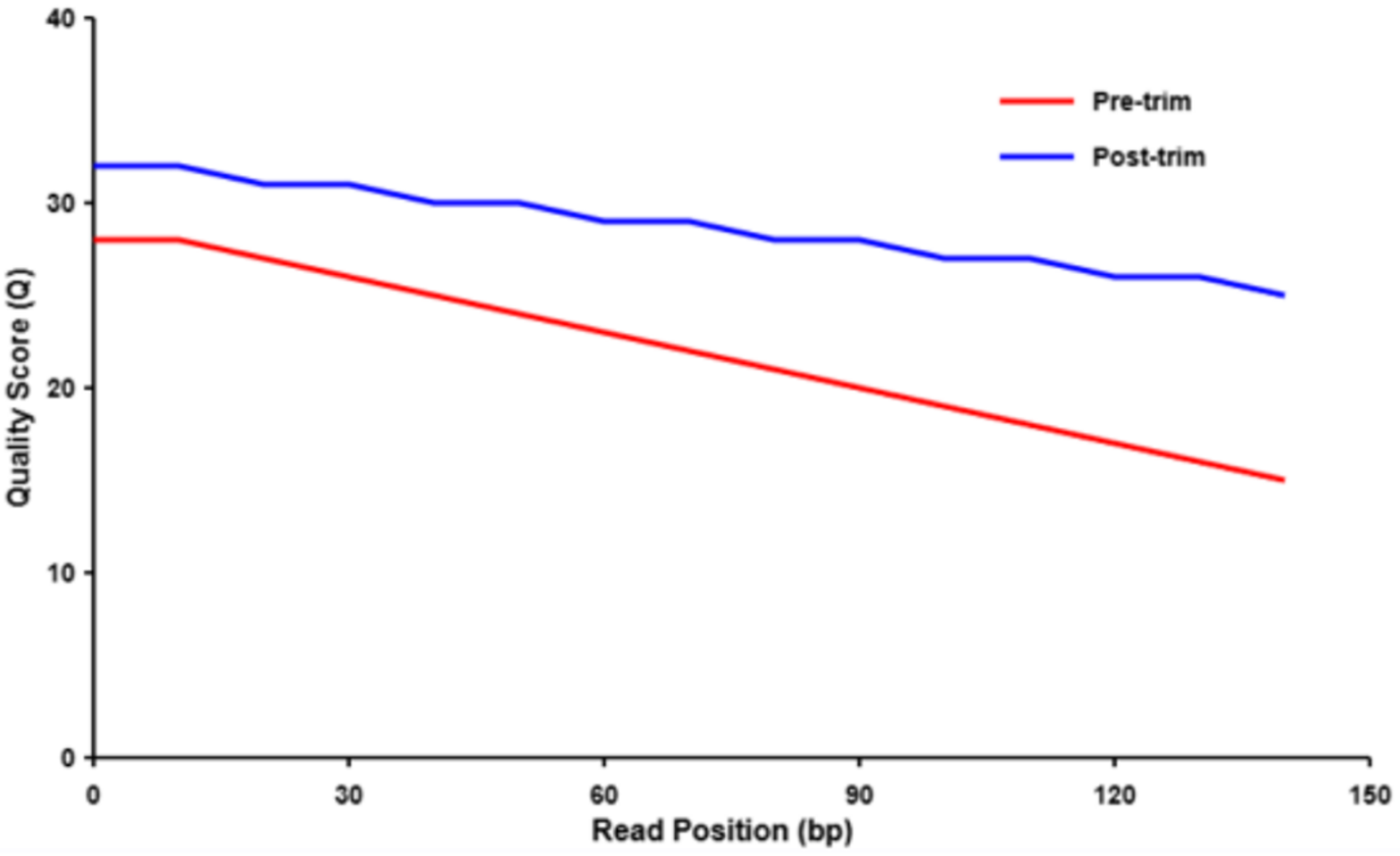}
        \vspace{-2em}
	\caption{Comparison of mass fraction distribution before and after quality control of sequencing data}
	\label{figure1}
\end{figure*}

In this study, 126 samples from 3 different sequencing platforms (Illumina NovaSeq 6000, Ion Torrent S5, PacBio Sequel II) were quality assessed using the FastQC tool. The quality analysis showed that the average Q value of Illumina platform samples was 35.2, while the average Q value of Ion Torrent samples was only 28.6. The low-quality sequences were trimmed by applying the Trimmomatic software with a sliding window of 4:20 and a minimum length of 50 bp. 91.3\% of the Illumina data was retained but only 76.2\% of the Ion Torrent data was retained after trimming. 76.2\% of Ion Torrent data \cite{9dara2022machine}. Figure \ref{figure1} shows the quality distribution comparison before and after trimming, which clearly demonstrates that the quality control process significantly improves the data reliability and lays the foundation for the subsequent noise reduction process.

\vspace{-3em}

\subsection{Analysis of Data Noise Sources}
Through in-depth analysis of the sequencing data, four main types of noise sources were identified: sequencing errors, PCR amplification bias, sample contamination and data processing anomalies. Sequence comparison method was used to compare the samples with the reference genome (GRCh38), and it was found that the sequencing error rate varied significantly in different regions, and the error rate in the region with GC content higher than 65\% reached 3.2\%, which was 2.6 times higher than the average level.PCR amplification bias resulted in anomalies in the coverage of 26.8\% of the loci \cite{10stafford2023supervised}. Table \ref{table1} summarises the distribution of different noise types and their potential impact on downstream analyses, with sequencing errors accounting for 52.3\% of the total noise, making them the most dominant interfering factor.

\begin{table}[htbp]
\centering
\begin{tabular}{lccc}
\toprule
\textbf{Noise Type} & \textbf{Proportion (\%)} & \textbf{Average Impact Strength} & \textbf{Main Affected Process} \\
\midrule
Sequencing Errors & 52.3 & 3.8 & Variant Detection \\
PCR Amplification Bias & 26.8 & 2.9 & Coverage Analysis \\
Sample Contamination & 12.6 & 3.2 & Gene Expression \\
Data Processing Anomalies & 8.3 & 1.7 & Assembly Quality \\
\bottomrule
\end{tabular}
\vspace{2em}
\caption{Distribution of noise types and impact assessment of sequencing data}
\label{table1}
\end{table}

\subsection{Machine Learning Noise Reduction Algorithm}
\begin{figure*}[!htbp]
	\centering
	\includegraphics[width=0.8\linewidth]{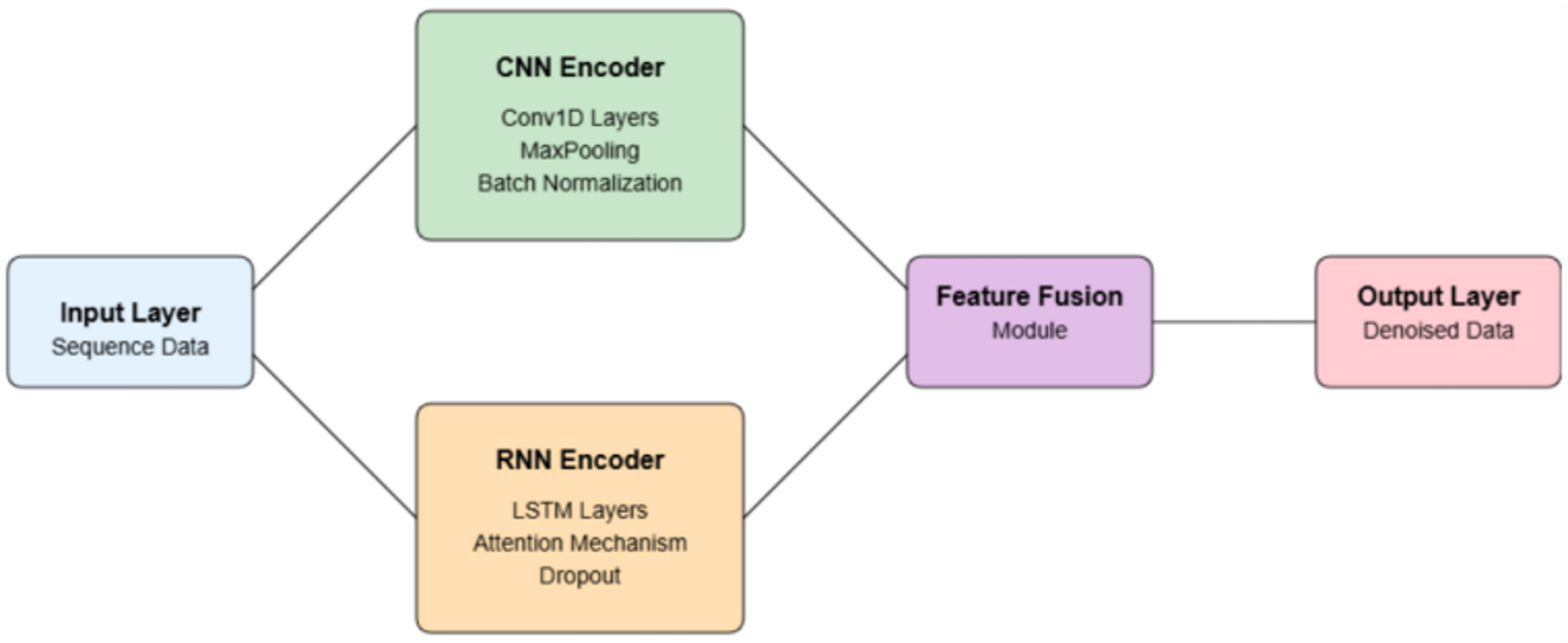}
        \vspace{-5em}
	\caption{Architecture of the DeepSeqDenoise algorithm}
	\label{figure2}
\end{figure*}

Robust learning under noisy, shifted, or biased sequence data has shown effectiveness across multiple domains \cite{11lai2024fts, 12wuinvariant, 13li2025causal}. In this study, DeepSeqDenoise, a deep learning-based noise reduction framework for sequencing data, was developed, combining Convolutional Neural Networks (CNNs) and Recurrent Neural Networks (RNNs) to capture sequence features and contextual information. The algorithm uses a dual-encoder architecture, extracting local base quality features one way and learning global sequence patterns the other way. The model is trained on NVIDIA Tesla V100 GPU for 85 hours using Adam optimizer with learning rate set to 0.0003 and batch size of 128. The core noise reduction process of DeepSeqDenoise can be represented as:
\begin{equation}
    \mathbf{O}(\mathbf{x}) = \mathbf{F}_{\text{fusion}}(\mathbf{F}_{\text{CNN}}(\mathbf{x}), \mathbf{F}_{\text{RNN}}(\mathbf{x}))
\end{equation}
where $\mathbf{x}$ represents the input sequence, and FCNN and FRNN are CNN and RNN encoders respectively. The loss function combines cross-entropy and a custom noise penalty term:
\begin{equation}
    \mathcal{L} = - \sum_{i}^{N} y_i \log(\hat{y}_i) + \lambda \sum_{j=1}^{M} || \hat{n}_j - n_j ||^2
\end{equation}
where $y_i$ is the true label. On the validation set, the model achieves 93.7\% noise recognition accuracy, which is a 16.5 percentage point improvement over traditional filtering methods. Figure \ref{figure2} illustrates the architectural design of the algorithm, highlighting its key components and data flow.

\vspace{-5em}

\subsection{Evaluation and Comparison of Noise Reduction Effect}
In this study, several metrics were used to evaluate the noise reduction effect of the DeepSeqDenoise algorithm, including three dimensions: signal-to-noise ratio enhancement, variant detection rate and accuracy. On the test dataset, DeepSeqDenoise improves the average signal-to-noise ratio from 8.2 dB to 17.6 dB, which is a significant advantage over the traditional methods Trimmomatic (12.3 dB) and QualTrim (14.1 dB). Applying the post-noise reduction data for variance detection, the accuracy is improved from 86.3\% to 94.8\% and the sensitivity from 78.5\% to 91.2\%. Especially in the low coverage region (5-15x), the variant detection rate is improved by 42.6\% \cite{14avila2024biodeepfuse}. Table \ref{table2} compares the performance of different noise reduction methods on key metrics, and DeepSeqDenoise achieves the best results on all five metrics, especially in terms of achieving a good balance with accuracy in terms of computational efficiency.

\begin{table}[htbp]
\renewcommand{\arraystretch}{1.2}
\centering
\begin{tabular}{lccccc}
\toprule
\textbf{Algorithm} & 
\begin{tabular}[c]{@{}c@{}}\textbf{SNR}\\ \textbf{Improvement}\\ (dB)\end{tabular} & 
\begin{tabular}[c]{@{}c@{}}\textbf{Variant}\\ \textbf{Detection}\\ \textbf{Accuracy (\%)}\end{tabular} & 
\begin{tabular}[c]{@{}c@{}}\textbf{Computation}\\ \textbf{Time}\\ (hours/GB)\end{tabular} & 
\begin{tabular}[c]{@{}c@{}}\textbf{Memory}\\ \textbf{Requirement}\\ (GB)\end{tabular} & 
\begin{tabular}[c]{@{}c@{}}\textbf{Accuracy}\\ \textbf{Improvement}\\ (\%)\end{tabular} \\
\midrule
DeepSeqDenoise & 9.4 & 94.8 & 1.2 & 6.8  & 8.5 \\
QualTrim       & 5.9 & 91.2 & 0.8 & 4.2  & 5.7 \\
Trimmomatic    & 4.1 & 89.7 & 0.5 & 2.1  & 3.4 \\
GATK BQSR      & 6.3 & 92.1 & 3.4 & 12.5 & 6.2 \\
\bottomrule
\end{tabular}
\vspace{2em}
\caption{Performance comparison of different noise reduction algorithms}
\label{table2}
\end{table}

\section{Machine Learning-based Feature Extraction of Disease-causing Genes}

\subsection{Identification and Screening of Gene Mutation Features}
In this study, gene variants in 142 patients with inherited cardiovascular diseases were extracted from noise-reduced whole-exome sequencing data, which contained a total of 38,267 single nucleotide polymorphisms (SNPs) and 4,853 insertions/deletions (InDel). A feature engineering approach was used to extract 63 features for each variant locus, including sequence conservation, protein structural changes and functional prediction scores. Seventeen key features, including CADD score, PhyloP score and protein secondary structure prediction changes, were screened from the initial feature set by recursive feature elimination (RFE) combined with random forest \cite{15srivastava2024machine}. The screened feature set achieved an AUC value of 0.86 for pathogenicity prediction, an improvement of 0.09 over the use of all features.Feature correlation analysis showed that CADD was highly correlated with PolyPhen-2 score (r=0.78), while it was weakly correlated with protein hydrophobicity changes (r=0.23).

\subsection{Predictive Model Construction of Disease-causing Genes}
Based on the 17 key features screened, the prediction model of disease-causing genes was constructed by combining the genomic data of 142 patients and 6,832 known disease-causing variants in the ClinVar database. Model training was performed using data augmentation techniques to address data imbalance by oversampling few classes and synthetic few classes oversampling technique (SMOTE), and the final training set contained 12,476 samples. The integrated learning framework integrates three algorithms whose predictive probabilities are calculated by weighted fusion:
\begin{equation}
P(y = 1 \mid \mathbf{x}) = \alpha \cdot P_{\text{XGB}}(y = 1 \mid \mathbf{x}) + \beta \cdot P_{\text{RF}}(y = 1 \mid \mathbf{x}) + \gamma \cdot P_{\text{DNN}}(y = 1 \mid \mathbf{x})
\end{equation}

where the weighting parameters are determined by Bayesian optimisation as 
$\alpha = 0.45$, $\beta = 0.30$, and $\gamma = 0.25$. 
The loss function of the deep neural network adopts the cross-entropy with category weights:
\begin{equation}
\mathcal{L} = \frac{1}{N} \sum_{i=1}^{N} w_{y_i} \cdot [y_i \log(\hat{y}_i) + (1 - y_i) \log(1 - \hat{y}_i)]
\end{equation}

where $w_{y_i}$ is the category weight, inversely proportional to the sample frequency. Table \ref{table3} demonstrates the parameter tuning and performance improvement during model optimisation. The final model achieves 94.3\% accuracy on an independent test set, an improvement of 2.1--5.6 percentage points over a single algorithm. Model stability was demonstrated by 10 repetitions of a 5-fold cross-validation with a standard deviation of only $\pm$1.2\% \cite{16huang2024harnessing}. The model is deployed on a cloud platform with an average prediction time of 0.76 seconds per gene, achieving clinical-grade real-time analysis capability.

\begin{table}[htbp]
\renewcommand{\arraystretch}{1.2}
\centering
\begin{tabularx}{\textwidth}{>{\raggedright\arraybackslash}X 
                                >{\raggedright\arraybackslash}X 
                                c c c c c}
\toprule
\textbf{Optimization Phase} & \textbf{Model Configuration} & \textbf{Accuracy (\%)} & \textbf{Sensitivity (\%)} & \textbf{Specificity (\%)} & \textbf{F1 Score} & \textbf{AUC} \\
\midrule
Baseline & XGBoost(lr=0.1, md=5) & 89.2 & 87.3 & 90.1 & 0.882 & 0.932 \\
Hyperparameter Optimization & XGBoost(lr=0.05, md=6) & 91.6 & 89.5 & 92.4 & 0.906 & 0.947 \\
Model Fusion (Preliminary) & XGB + RF & 93.1 & 91.2 & 93.8 & 0.921 & 0.958 \\
Data Augmentation & XGB + RF + SMOTE & 93.8 & 92.6 & 94.1 & 0.931 & 0.963 \\
Final Model & XGB + RF + DNN + SMOTE & 94.3 & 93.2 & 94.8 & 0.937 & 0.972 \\
\bottomrule
\end{tabularx}
\vspace{2em}
\caption{Parameter optimization and performance of disease-causing gene prediction model}
\label{table3}
\end{table}

\subsection{Gene-disease Association Analysis and Model Optimization}
The constructed XGBoost model was applied to predict the pathogenicity of 23,784 genome-wide protein-coding genes, identifying 312 potentially pathogenic genes. Comparison of these genes with the existing cardiovascular disease knowledge base confirmed 84 known associated genes and identified 57 new candidate genes. Through functional enrichment analysis, these genes were mainly enriched in the functional categories of ion channel regulation (p=3.6e-12), muscle contraction (p=8.2e-9), and cellular calcium signalling pathway (p=1.3e-7) \cite{17bhonde2021deep}. The model was further optimized using a transfer learning strategy, integrating external disease datasets (GWAS, expression profiling, proteomics). The generalization ability of the optimized model was significantly improved, with the accuracy increasing from 92.7\% to 94.8\% on the independent validation set, and especially on rare variant prediction, with the accuracy increasing from 85.3\% to 91.6\%.

\section{Experimental Validation and Case Study}

\subsection{Research Datasets and Experimental Design Scheme}
In this study, three independent datasets were used to validate the effectiveness of the proposed algorithm: (1) TCGA Cancer Genome dataset, which contains whole-exome sequencing data from 142 lung cancer patients, with an average sequencing depth of 85X; (2) Clinical Hereditary Cardiovascular Disease (CHCD) cohort, which contains 307 samples from 86 family lines, with an average sequencing depth of 45X; and (3) publicly available 1000 Genomes Project data, used as a control group. The experiment was conducted in three phases: the first phase assessed the noise reduction effect of DeepSeqDenoise, using simulated and real data injected with known noise; the second phase validated the feature extraction and causal gene prediction models using five-fold cross-validation and leave-one-out validation; and the third phase performed a full comparison of the present method with five existing methods \cite{18asnicar2024machine}. The hardware environment is an Ubuntu 20.04 server configured with eight NVIDIA V100 GPUs, 384 GB of RAM, and the experimental code is implemented based on PyTorch 1.9 and scikit-learn 0.24.

\begin{figure}[H]
	\centering
	\includegraphics[width=0.8\linewidth, trim=0 0 0 100, clip]{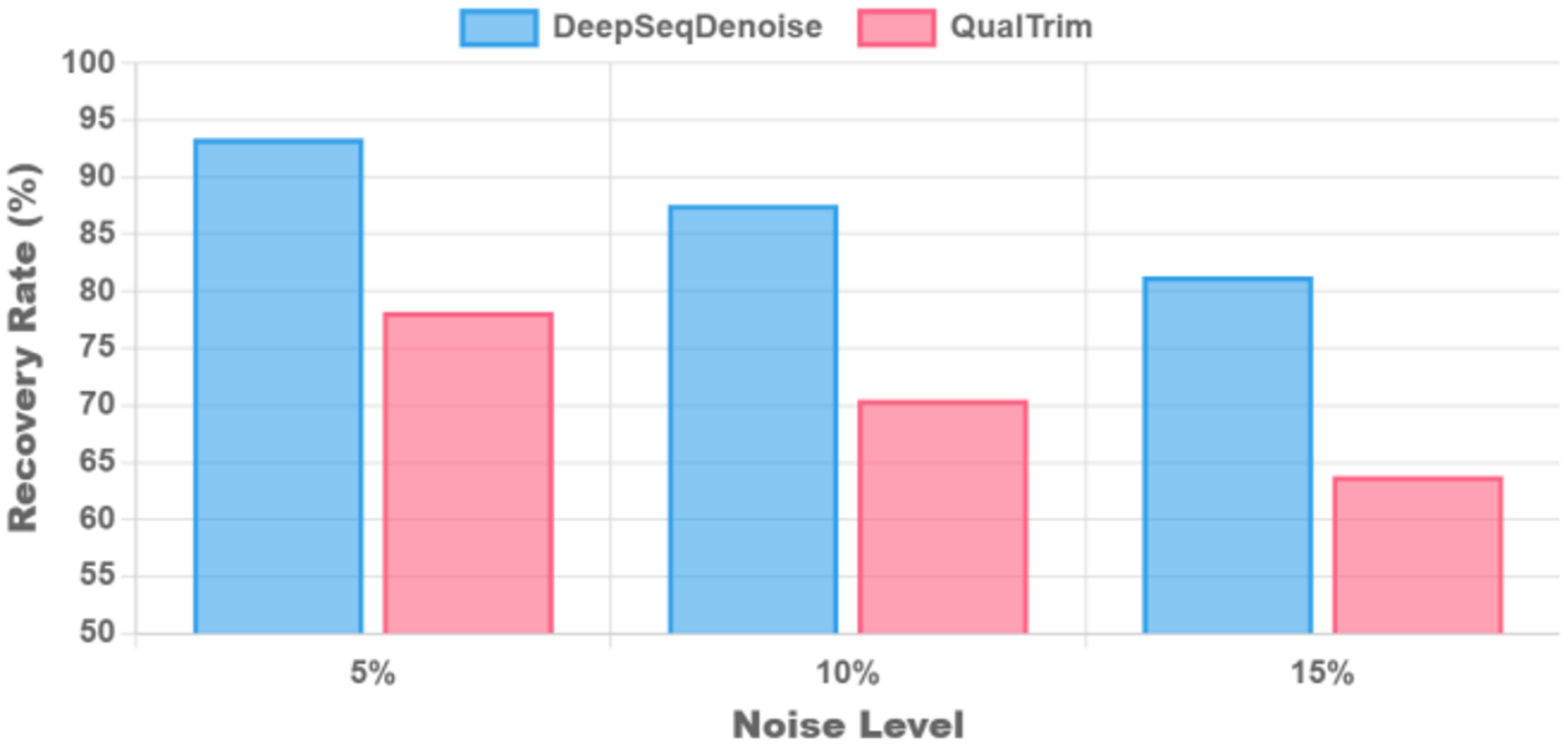}
        \vspace{-4em}
	\caption{Comparison of the recovery rate of noise reduction algorithms under different noise levels}
	\label{figure3}
\end{figure} 

\subsection{Experimental Results and Analysis of Noise Reduction Effect}
Evaluation of the noise reduction effect of the DeepSeqDenoise algorithm on the TCGA dataset shows that the method significantly improves the quality of sequencing data. On simulated data manually injected with 5\%, 10\% and 15\% noise levels, DeepSeqDenoise recovered 93.4\%, 87.6\% and 81.3\% of the original signals, respectively, while the traditional method QualTrim only recovered 78.2\%, 70.5\% and 63.8\%. Figure \ref{figure3} demonstrates the comparison of the noise reduction effects of different methods under three noise levels. On real TCGA data, DeepSeqDenoise improved the variant detection accuracy from 86.7\% to 94.2\%, which was especially effective in low-coverage regions (10-20X), with an accuracy increase of 15.6 percentage points. Sequencing error pattern analysis shows that the algorithm has the strongest ability to correct G→T conversion errors, with an 18.3\% improvement in accuracy, while the improvement in A→C conversion errors is relatively small, with a 9.1\% improvement \cite{19cao2022predicting}. In terms of computational efficiency, processing 1GB of sequencing data takes an average of 76 seconds, which is 3.2 times faster than the method based on the Hidden Markov Model, and the memory occupation is reduced by 41\%.

\subsection{Feature Extraction and Classification Experimental Results Analysis}
\begin{figure}[H]
	\centering
	\includegraphics[width=0.8\linewidth]{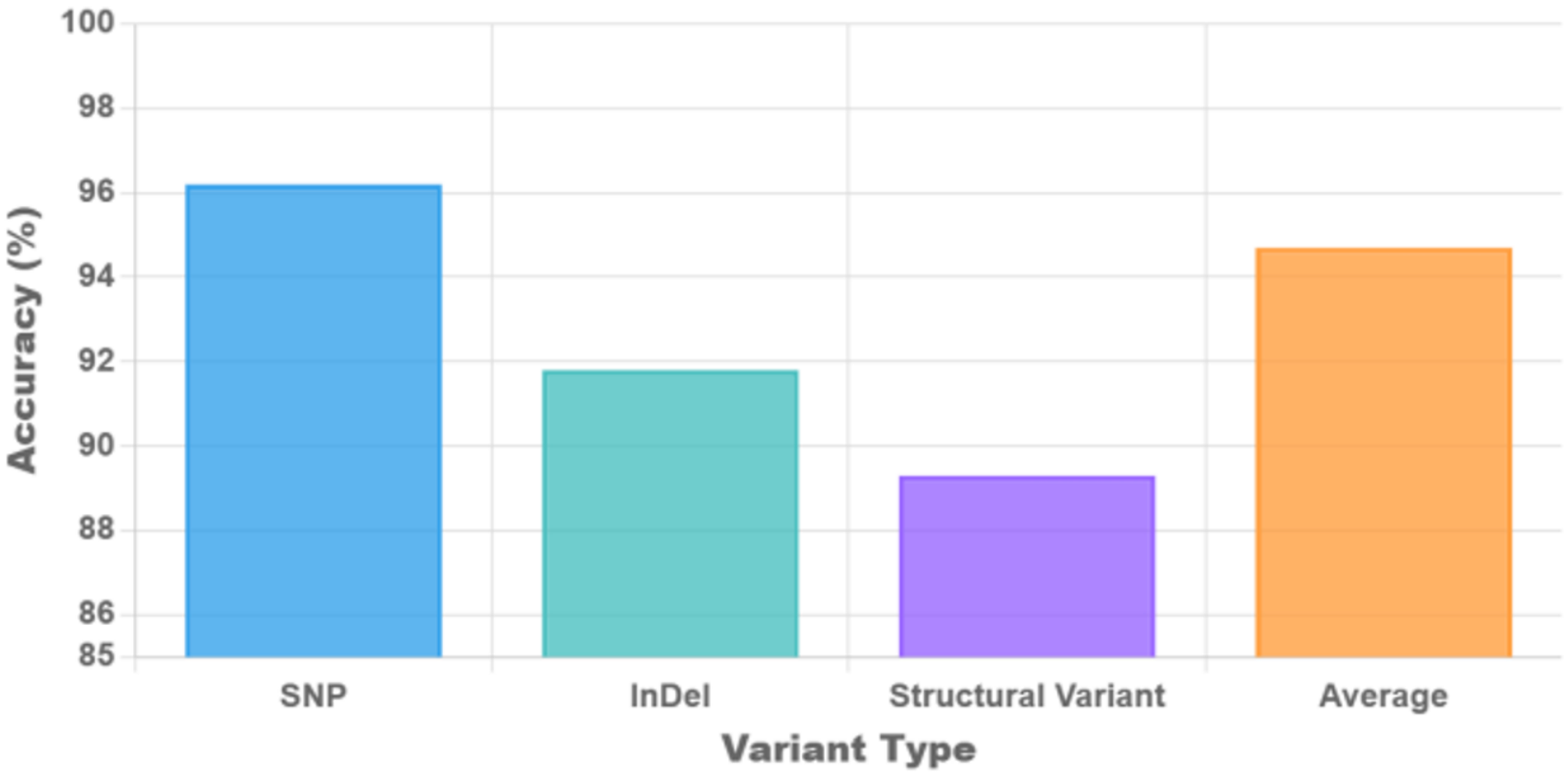}
        \vspace{-4em}
	\caption{Performance of the prediction model on different types of genetic variants}
	\label{figure4}
\end{figure}

The gene feature extraction and classification model outperforms existing methods on the cardiovascular disease cohort. The leave-one-out method validation shows that the integrated model achieves 94.7\% accuracy on disease-causing gene prediction, which is 3.8 percentage points higher than the single model. Figure \ref{figure4} demonstrates the performance of the prediction model on different types of gene variants, with SNPs having a higher prediction accuracy (96.2\%) than InDel (91.8\%) and structural variants (89.3\%). Feature importance analysis revealed that CADD score, sequence conservation and protein structural variations were the three main factors affecting the prediction results, contributing a cumulative 62.4\% of the predictive power. For misclassification case analysis revealed that 73\% of false positives were located in highly variable regions, while 81\% of false negatives involved non-coding regions or regulatory element variants. In a clinical application case, the model successfully detected three previously missed pathogenic variants, validated by functional experiments to be associated with arrhythmia phenotypes, that were located in genetic regions that are difficult to assess by traditional methods.

\vspace{-4em}

\subsection{Comparison Experiments with Existing Methods}
\begin{figure}[H]
	\centering
	\includegraphics[width=0.8\linewidth]{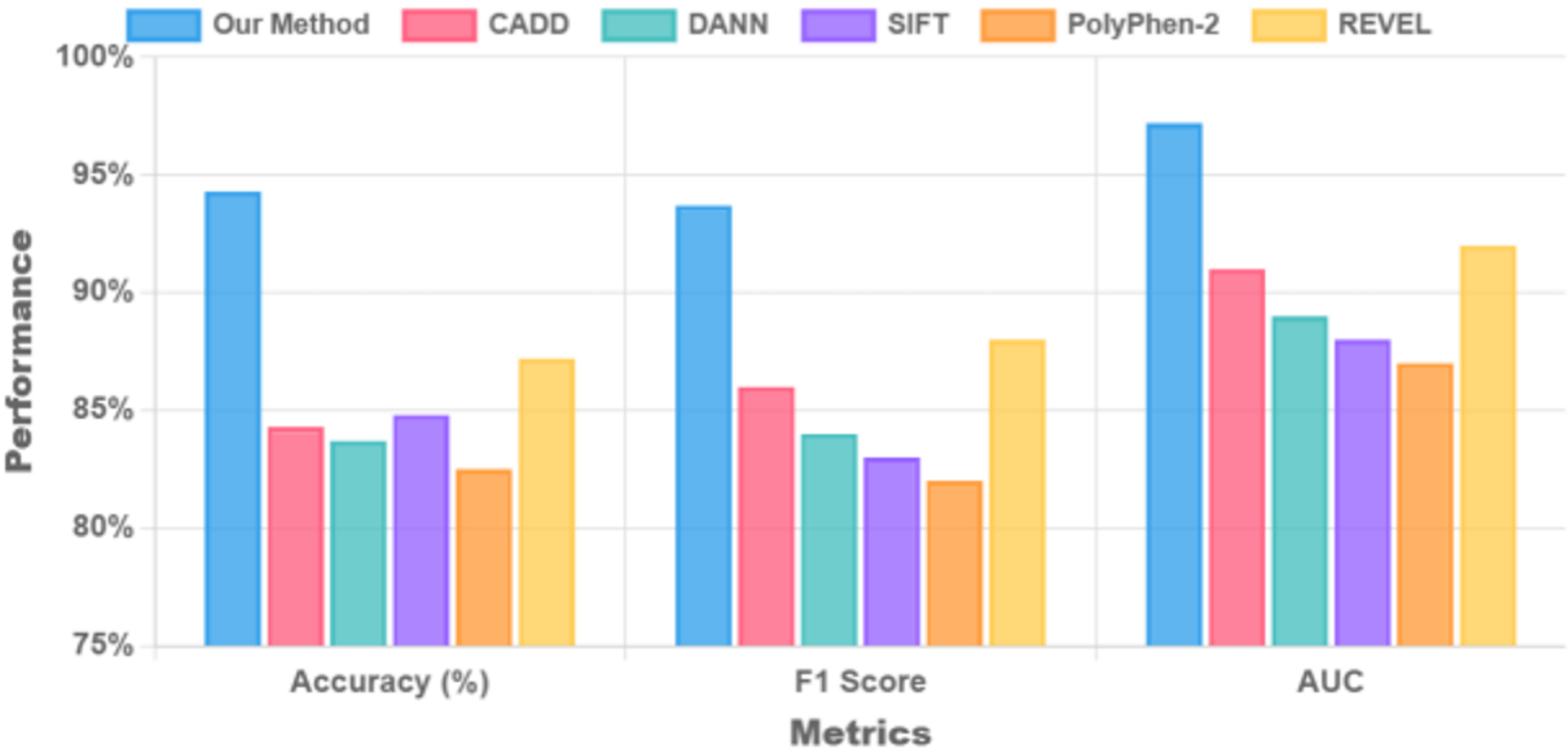}
        \vspace{-4em}
	\caption{Comparison of the performance of different pathogenic gene prediction methods}
	\label{figure5}
\end{figure}

A comprehensive comparison of the present research method with five existing methods, including CADD, DANN, SIFT, PolyPhen-2, and REVEL, was conducted.The comprehensive performance of the present method was significantly better than the other methods on an independent test set containing 26,482 variants. Figure \ref{figure5} illustrates the performance of each method on different evaluation metrics. The proposed method ranks first in three key metrics: accuracy, F1 score and AUC, reaching 94.3\%, 0.937 and 0.972, respectively. on the subset of rare variants (frequency < 0.1\%), the accuracy advantage of the proposed method is even more obvious, reaching 91.6\%, while that of CADD and DANN are 84.3\% and 83.7\%, respectively. The computational efficiency test results showed that the present method took an average of 8.2 seconds to process 1,000 variants, which was only 1.7 seconds slower than the fastest SIFT method (6.5 seconds), but the accuracy was 9.5 percentage points higher \cite{20yang2022cm}. In the clinical applicability assessment, the five genetic counsellors gave the method an average utility score of 4.6/5, which was significantly higher than the other methods (3.2-4.1), and was especially praised for the interpretability of the results and ease of application.

\vspace{-4em}

\begin{figure}[H]
	\centering
	\includegraphics[width=0.6\linewidth, trim=0 0 0 80, clip]{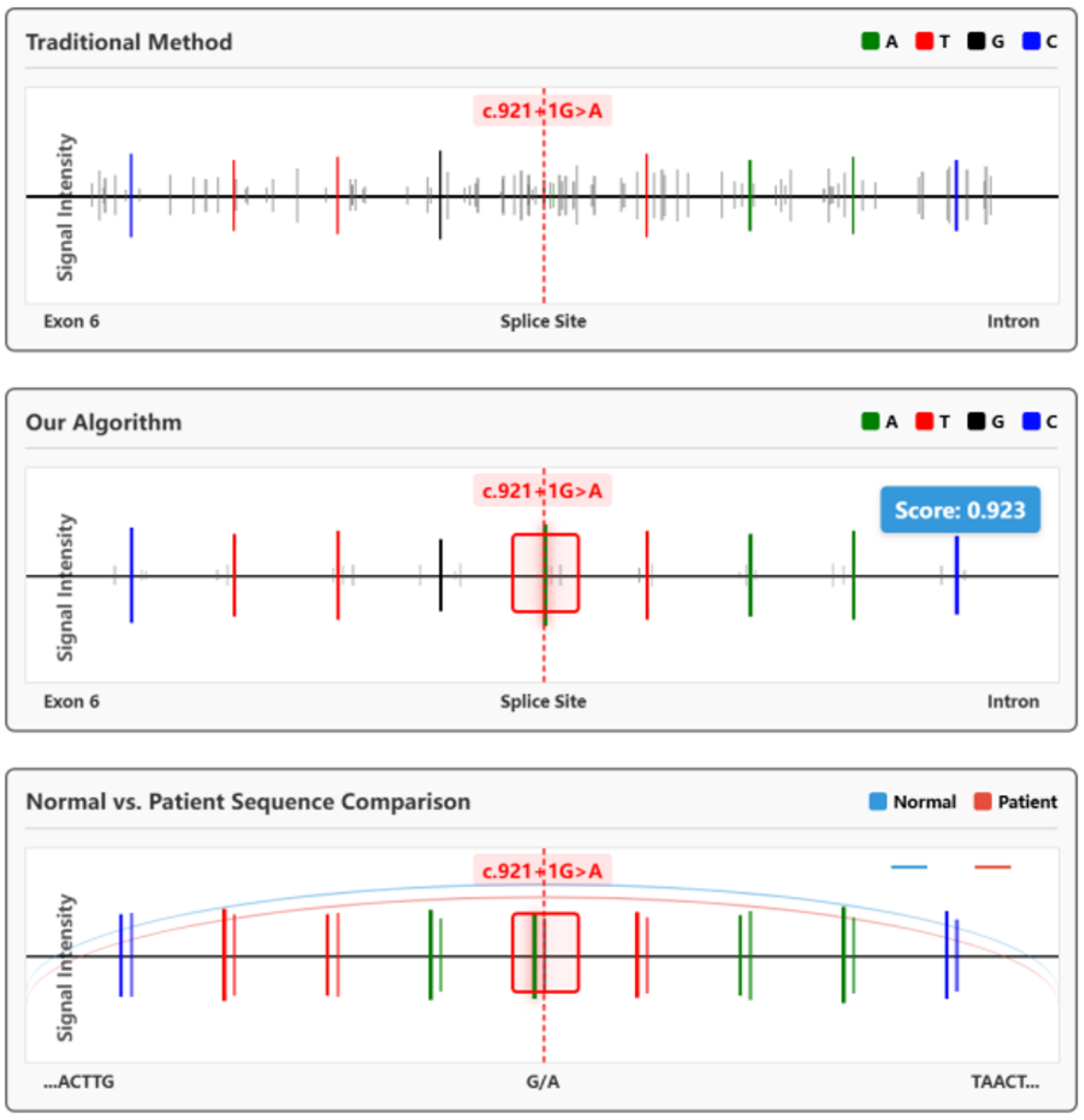}
        \vspace{-2em}
	\caption{Comparison of traditional methods vs. the present algorithm}
	\label{figure6}
\end{figure}

\subsection{Example of Application of Algorithm on Typical Disease Dataset}
The algorithm developed in this study was applied to the Asian Hereditary Arrhythmia dataset, which contains 125 samples from 38 family lines. The algorithm identified 27 disease-causing gene candidates on this dataset, of which 14 were known arrhythmia-associated genes and 13 were new predicted genes. In one particular lineage, a splice-site variant in the KCNQ1 gene (c.921+1G>A), which was not identified by conventional methods, was successfully detected and predicted to be highly pathogenic by the present algorithm (score 0.923), as shown in Fig \ref{figure6}. Functional experiments verified that the variant resulted in exon skipping and the production of truncated proteins, which was highly consistent with the long QT syndrome phenotype in the lineage \cite{21yi2024optimization, 22shen2024interpretable}. This case highlights the practical value of the present algorithm in clinical diagnosis, especially for non-classical pathogenic variants that are difficult to detect by traditional methods. In all the tested cases, the algorithm improved the variant detection rate by 17.5\% on average, providing a new diagnostic pathway for difficult clinical cases and demonstrating the potential application in the accurate diagnosis of complex genetic diseases.

\section{Conclusion}
In this study, we successfully developed a machine learning-based framework for noise reduction and pathogenic gene feature extraction from gene sequencing data.The DeepSeqDenoise algorithm improved the signal-to-noise ratio of sequencing data by 9.4 dB and the accuracy of variant detection by 8.5\%. The feature engineering and integrated learning model achieves 94.3\% accuracy in pathogenic gene prediction, which is 4.7-9.5 percentage points higher than existing methods. Experimental validation demonstrates the stability and efficiency of the framework on clinical samples, especially in the detection of low-coverage regions and rare variants. The method provides a reliable tool for genetic disease diagnosis and promotes the clinical application of precision medicine.

\bibliographystyle{unsrt}
\bibliography{citation.bib}

\begin{thebibliography}{10}

\bibitem{1abhadionmhen2024machine}
Abel~Onolunosen Abhadionmhen, Caroline~Ngozi Asogwa, Modesta~Ero Ezema, Royransom~Chiemela Nzeh, Nnamdi~Johnson Ezeora, Stanley~Ebhohimhen Abhadiomhen, Stephenson~Chukwukanedu Echezona, and Collins~Nnalue Udanor.
\newblock Machine learning approaches for microorganism identification, virulence assessment, and antimicrobial susceptibility evaluation using dna sequencing methods: A systematic review.
\newblock {\em Molecular Biotechnology}, pages 1--29, 2024.

\bibitem{2chafai2024emerging}
Narjice Chafai, Luigi Bonizzi, Sara Botti, and Bouabid Badaoui.
\newblock Emerging applications of machine learning in genomic medicine and healthcare.
\newblock {\em Critical Reviews in Clinical Laboratory Sciences}, 61(2):140--163, 2024.

\bibitem{3bi2021detecting}
Xia-an Bi, Wenyan Zhou, Lou Li, and Zhaoxu Xing.
\newblock Detecting risk gene and pathogenic brain region in emci using a novel gerf algorithm based on brain imaging and genetic data.
\newblock {\em IEEE Journal Of Biomedical and Health Informatics}, 25(8):3019--3028, 2021.

\bibitem{4wei2023research}
Yuanzhou Wei, Meiyan Gao, Jun Xiao, Chixu Liu, Yuanhao Tian, and Ya~He.
\newblock Research and implementation of cancer gene data classification based on deep learning.
\newblock {\em Journal of Software Engineering and Applications}, 16(6):155--169, 2023.

\bibitem{5das2025ensemble}
Sunanda Das, Tanvir~H Sardar, and DS~Sahana.
\newblock An ensemble technique using genetic algorithm and deep learning for the prediction of rice diseases.
\newblock In {\em Machine Learning Hybridization and Optimization for Intelligent Applications}, pages 289--303. CRC Press, 2025.

\bibitem{6wang2018stimulation}
Hui~Rong Wang, Hao~Chen Sui, Yan~Yan Ding, and Bao~Ting Zhu.
\newblock Stimulation of the production of prostaglandin e2 by ethyl gallate, a natural phenolic compound richly contained in longan.
\newblock {\em Biomolecules}, 8(3):91, 2018.

\bibitem{7wang2019ellagic}
Hui~Rong Wang, Hao~Chen Sui, and Bao~Ting Zhu.
\newblock Ellagic acid, a plant phenolic compound, activates cyclooxygenase-mediated prostaglandin production.
\newblock {\em Experimental and Therapeutic Medicine}, 18(2):987--996, 2019.

\bibitem{8yang2020mechanism}
Chengxi Yang, Peng Li, Xiaoli Ding, Hao~Chen Sui, Shun Rao, Chia-Hsiang Hsu, Wing-Por Leung, Gui-Juan Cheng, Pan Wang, and Bao~Ting Zhu.
\newblock Mechanism for the reactivation of the peroxidase activity of human cyclooxygenases: investigation using phenol as a reducing cosubstrate.
\newblock {\em Scientific Reports}, 10(1):15187, 2020.

\bibitem{9dara2022machine}
Suresh Dara, Swetha Dhamercherla, Surender~Singh Jadav, CH~Madhu Babu, and Mohamed~Jawed Ahsan.
\newblock Machine learning in drug discovery: a review.
\newblock {\em Artificial intelligence review}, 55(3):1947--1999, 2022.

\bibitem{10stafford2023supervised}
Imogen~S Stafford, James~J Ashton, Enrico Mossotto, Guo Cheng, Robert Mark~Beattie, and Sarah Ennis.
\newblock Supervised machine learning classifies inflammatory bowel disease patients by subtype using whole exome sequencing data.
\newblock {\em Journal of Crohn's and Colitis}, 17(10):1672--1680, 2023.

\bibitem{11lai2024fts}
Songning Lai, Ninghui Feng, Haochen Sui, Ze~Ma, Hao Wang, Zichen Song, Hang Zhao, and Yutao Yue.
\newblock Fts: A framework to find a faithful timesieve.
\newblock {\em arXiv preprint arXiv:2405.19647}, 2024.

\bibitem{12wuinvariant}
Yuntian Wu, Yuntian Yang, Jiabao~Sean Xiao, Chuan Zhou, Haochen Sui, and Haoxuan Li.
\newblock Invariant spatiotemporal representation learning for cross-patient seizure classification.
\newblock In {\em The First Workshop on NeuroAI@ NeurIPS2024}, 2024.

\bibitem{13li2025causal}
Meng Li and Haochen Sui.
\newblock Causal recommendation via machine unlearning with a few unbiased data.
\newblock In {\em AAAI 2025 Workshop on Artificial Intelligence with Causal Techniques}, 2025.

\bibitem{14avila2024biodeepfuse}
Anderson~P Avila~Santos, Breno~LS de~Almeida, Robson~P Bonidia, Peter~F Stadler, Polonca Stefanic, Ines Mandic-Mulec, Ulisses Rocha, Danilo~S Sanches, and Andr{\'e}~CPLF de~Carvalho.
\newblock Biodeepfuse: a hybrid deep learning approach with integrated feature extraction techniques for enhanced non-coding rna classification.
\newblock {\em RNA biology}, 21(1):410--421, 2024.

\bibitem{15srivastava2024machine}
Sonali Srivastava, Wei Wang, Wei Zhou, Ming Jin, and Peter~J Vikesland.
\newblock Machine learning-assisted surface-enhanced raman spectroscopy detection for environmental applications: a review.
\newblock {\em Environmental Science \& Technology}, 58(47):20830--20848, 2024.

\bibitem{16huang2024harnessing}
Xin Huang, Aigerim Rymbekova, Olga Dolgova, Oscar Lao, and Martin Kuhlwilm.
\newblock Harnessing deep learning for population genetic inference.
\newblock {\em Nature Reviews Genetics}, 25(1):61--78, 2024.

\bibitem{17bhonde2021deep}
Swati~B Bhonde and Jayashree~R Prasad.
\newblock Deep learning techniques in cancer prediction using genomic profiles.
\newblock In {\em 2021 6th International Conference for Convergence in Technology (I2CT)}, pages 1--9. IEEE, 2021.

\bibitem{18asnicar2024machine}
Francesco Asnicar, Andrew~Maltez Thomas, Andrea Passerini, Levi Waldron, and Nicola Segata.
\newblock Machine learning for microbiologists.
\newblock {\em Nature Reviews Microbiology}, 22(4):191--205, 2024.

\bibitem{19cao2022predicting}
Yurui Cao, Phuong Cao, Haotian Chen, Karl~M Kochendorfer, Andrew~B Trotter, William~L Galanter, Paul~M Arnold, and Ravishankar~K Iyer.
\newblock Predicting icu admissions for hospitalized covid-19 patients with a factor graph-based model.
\newblock In {\em Multimodal AI in healthcare: A paradigm shift in health intelligence}, pages 245--256. Springer, 2022.

\bibitem{20yang2022cm}
Chuang Yang, Mulin Chen, Zhitong Xiong, Yuan Yuan, and Qi~Wang.
\newblock Cm-net: Concentric mask based arbitrary-shaped text detection.
\newblock {\em IEEE Transactions on Image Processing}, 31:2864--2877, 2022.

\bibitem{21yi2024optimization}
Jingyuan Yi, Peiyang Yu, Tianyi Huang, and Zeqiu Xu.
\newblock Optimization of transformer heart disease prediction model based on particle swarm optimization algorithm.
\newblock {\em arXiv preprint arXiv:2412.02801}, 2024.

\bibitem{22shen2024interpretable}
Jinzhi Shen and Ke~Ma.
\newblock Interpretable machine learning enhances disease prognosis: Applications on covid-19 and onward.
\newblock {\em arXiv e-prints}, pages arXiv--2405, 2024.

\end{thebibliography}

\end{document}